\documentclass{article}
\usepackage[utf8]{inputenc}

\usepackage[normalem]{ulem}

\usepackage{nicematrix}

\usepackage{tikz}
\usetikzlibrary{positioning}

\usepackage[a4paper,top=2cm,bottom=2cm,left=2.5cm,right=2.5cm,marginparwidth=2cm]{geometry}

\usepackage{hyperref}
\hypersetup{
    colorlinks=true,
    linkcolor=blue,
    filecolor=magenta,      
    urlcolor=cyan,
    pdftitle={Overleaf Example},
    pdfpagemode=FullScreen,
    }

\usepackage{algpseudocode}

\usepackage{hyperref}
\usepackage{array}
\usepackage{mdframed}
\usepackage{mathtools}

\usepackage{verbatim}

\usepackage{tikz}
\usetikzlibrary{positioning}
\usetikzlibrary{arrows}
\usetikzlibrary{shapes.multipart}

\usepackage{booktabs}

\usepackage{amsthm}
\usepackage{amssymb}
\usepackage{amsmath, ulem}
\usepackage{blkarray}
\usepackage{graphicx}
\usepackage{multicol}

\usepackage{xcolor}

\usepackage{caption}
\usepackage{subcaption}

\newcommand{\stkout}[1]{\ifmmode\text{\sout{\ensuremath{#1}}}\else\sout{#1}\fi}

\usepackage{multirow}
\usepackage{soul}

\usepackage{textcomp}

\usepackage{accents}

\usepackage{algorithm}

\DeclareMathOperator*{\argmin}{argmin}
\DeclareMathOperator*{\argmax}{argmax}

\newtheorem{theorem}{Theorem}
\newtheorem{lemma}{Lemma}
\newtheorem{proposition}{Proposition}

\usepackage{csquotes}

\newcommand{\suchthat}{\;\ifnum\currentgrouptype=16 \middle\fi|\;}

\newcommand\R{\mathbb{R}}

\newcommand\calG{\mathcal G}
\newcommand\calH{\mathcal H}

\newcommand\tto{\rightrightarrows}

\newcommand{\gph}{\operatorname{gph}}

\DeclareMathAlphabet{\mathpzc}{OT1}{pzc}{m}{it}

\newcounter{casenum}
\newenvironment{caseof}{\setcounter{casenum}{1}}{\vskip.5\baselineskip}
\newcommand{\case}[2]{\vskip.5\baselineskip\par\noindent {\bfseries Case \arabic{casenum}:} #1\\#2\addtocounter{casenum}{1}}

\definecolor{mygreen}{rgb}{0.0,0.7,0.0}
\definecolor{mybrown}{rgb}{0.5,0.5,0.0}

\title{
{\huge \textbf{Adversarial training with restricted data manipulation}}}
\author{\\[1ex]
David Benfield$^{a}$, Stefano Coniglio$^{b}$, Phan Tu Vuong$^{a}$, and Alain Zemkoho$^{a}$\\[1.5ex]
$^a$School of Mathematical Sciences\\ University of Southampton\\  SO17 1BJ Southampton, United Kingdom \\
\href{db3g17@soton.ac.uk}{db3g17@soton.ac.uk}, \href{t.v.phan@soton.ac.uk}{t.v.phan@soton.ac.uk}, \href{a.b.zemkoho@soton.ac.uk}{a.b.zemkoho@soton.ac.uk}\\[1.5ex]
$^b$Department of Economic Sciences\\
University of Bergamo \\
24127 Bergamo, Italy\\
\href{stefano.coniglio@unibg.it}{stefano.coniglio@unibg.it}}
\date{}

\begin{document}


\maketitle


\begin{abstract}
Adversarial machine learning concerns situations in which learners face attacks from active adversaries. Such scenarios arise in applications such as spam email filtering, malware detection and fake-image generation, where security methods must be actively updated to keep up with the ever-improving generation of malicious data. Pessimistic Bilevel optimisation has been shown to be an effective method of training resilient classifiers against such adversaries. By modelling these scenarios as a game between the learner and the adversary, we anticipate how the adversary will modify their data and then train a resilient classifier accordingly. However, since existing pessimistic bilevel approaches feature an unrestricted adversary, the model is vulnerable to becoming overly pessimistic and unrealistic. When finding the optimal solution that defeats the classifier, it's possible that the adversary's data becomes nonsensical and loses its intended nature. Such an adversary will not properly reflect reality, and consequently, will lead to poor classifier performance when implemented on real-world data. By constructing a constrained pessimistic bilevel optimisation model, we restrict the adversary's movements and identify a solution that better reflects reality. We demonstrate through experiments that this model performs, on average, better than the existing approach.
\end{abstract}



\section{Introduction}
\label{sctn:Introduction}

Adversarial machine learning considers the exploitable vulnerabilities of machine learning models and the strategies needed to counter or mitigate such threats \cite{adv_book_2023}. By considering these vulnerabilities during the development stage of our machine learning models, we can work to build resilient methods \cite{adv_deep_learning_survey, GameTheorySurvey} such as protection from credit card fraud \cite{INFORMS_credit_card_fraud_2} or finding the optimal placement of air defence systems \cite{INFORMS_defense}. In particular, we consider the model's sensitivity to changes in the distribution of the data. The way the adversary influences the distribution can fall under numerous categories, see \cite{taxonomy} for a helpful taxonomy that categorises these attacks. We focus on the specific case of \textit{exploratory} attacks, which consider the scenarios where adversaries attempt to modify their data to evade detection by a classifier. Such attacks might occur in security scenarios such as malware detection \cite{Malware_Detection} and network intrusion traffic \cite{Network_Intrusion}. In a similar vein, and more recently, vulnerabilities in deep neural networks (DNN) are being discovered, particularly in the field of computer vision and image classification; small perturbations in the data can lead to incorrect classifications by the DNN \cite{DNNAttacks1, DNNAttacks2}. These vulnerabilities raise concerns about the robustness of the machine learning technology that is being adopted and, in some cases, in how safe relying on their predictions could be in high-risk scenarios such as autonomous driving \cite{DNNSelfDriving} and medical diagnosis \cite{DNNMedical}.
By modelling the adversary's behaviour and anticipating these attacks, we can train classifiers that are resilient to such changes in the distribution before they occur.

Game theoretic models have been demonstrated to be an effective method of modelling these attacks. These models see one player, the learner, attempt to train a classifier while another player, the adversary, modifies their data in an attempt to evade detection by the learner's classifier, see for example \cite{Adv_Class} for early work in modelling adversarial learning. The precise structure of such a model then depends on a number of factors, such as whether the attack occurs either at training time \cite{Bruck_Ext}, or implementation time \cite{Brückner_Scheffer_2011}. Further to this, while some games assume the players act simultaneously \cite{NE_static, Deep_learning_Games}, others allow for sequential play where either the adversary acts first \cite{Liuetall, AdvLeadExt, AdvLeadExtTwo, kantarcıoğlu_xi_clifton_2010} or the learner acts first \cite{taxonomy, Brückner_Scheffer_2011}. We focus our attention on attacks made at implementation time where the adversary seeks to evade an already established classifier. Indeed, existing game theoretic approaches, e.g. \cite{Bruck_Ext, bruck_Ext_2_1, benfield2024classificationstrategicadversarymanipulation}, have demonstrated, through experiments, the potential to yield improved classifier performance on adversarial data. Pessimistic bilevel optimisation has proved to be a particularly promising approach to modelling these scenarios due to its ability to capture the antagonistic nature of the adversary when multiple optimal strategies are available to them \cite{Brückner_Scheffer_2011, benfield2024classificationstrategicadversarymanipulation}. However, within these approaches, the adversary's only task is to produce data that evades detection while not necessarily considering the quality of the data produced. Such models can allow for the creation of a perfect adversary that produces instances of adversarial data that are indistinguishable from legitimate data. Such results are expected, and in fact desired, in some applications. For example, the purpose of a Generative Adversarial Network (GAN) is to generate legitimate images where the ultimate aim to produce images that perfectly replicate the legitimate. However, in the case of detecting spam emails or fraudulent bank transactions, it is unlikely that the adversarial data perfectly replicates the legitimate. If a spam email is, in every way, identical to a legitimate email, then it will lose its intended message. Consequently, the data generated in the bilevel model does not adequately reflect the real world but rather over-estimates the ability of the adversary. The resulting classifier will hence suffer in performance \cite{benfield2024classificationstrategicadversarymanipulation}.


Motivating the adversary towards particular movements has been shown to be an effective strategy. In the pessimistic bilevel setting, the authors in \cite{Brückner_Scheffer_2011} introduce an $l_2$ regularisation on the distance that the adversary moves their data. In this way, the adversary must find a solution that is close to its original position. Numerical experiments demonstrated the effectiveness of such a model, however, their proposed solution method required strong convexity which restricted the choices of regularisation. In some scenarios, such as text-based classification, where the data are represented by some embedding, there might be more appropriate choices to measure how far the adversary has moved their data. Google's BERT \cite{BERT}, for example, embeds the data by measuring the cosine similarity between words. In this case, it might instead be more appropriate to also use the cosine similarity to measure the distance that the adversary has moved their data. However, since the cosine similarity is non-convex, it cannot be used in this model. Therefore, in this work, we present a novel pessimistic bilevel model with lower-level constraints that implement restrictions on the movement of the adversary's data. By making no assumptions on the convexity of these constraint functions, we can present a general model that allows for any continuously differentiable choice of similarity measure.

In summary, the contributions in this paper include the construction of a novel contained pessimistic bilevel program to model adversarial scenarios. We make no assumptions on the convexity of the lower-level problem or uniqueness of its solution as was similarly done with the unconstrained model in \cite{benfield2024classificationstrategicadversarymanipulation}. However, by introducing constraints how far the adversary can move their data, we create a model that more consistently trains a high-performing classifier. Measurement of adversarial movement in a pessimistic bilevel model has previously been investigated in \cite{Brückner_Scheffer_2011}. However, due to strong assumptions about the convexity of the lower-level, this approach is limited to its choices of how this movement is measured. We present a model that makes no assumptions on the convexity of the lower-level problem which allows more freedom over this choice of measurement. This allows us to use a more appropriate choice for text-based classification which, through numerical experiments, is shown to perform better.

The outline of the paper is as follows, in Section \ref{sctn:bilevelModel} we construct a novel constrained pessimistic bilevel program to model adversarial attacks, and subsequently, we demonstrate how the model can be applied to text-based classification tasks and study its properties under this application. Then, in section \ref{sctn:solution method}, we outline the solution method for solving the bilevel problem before demonstrating though experiments in section \ref{sctn:Experiments}, that our model performs more consistently than the existing approaches.

\section{The mathematical model}
\label{sctn:bilevelModel}

We model test-time adversarial attacks as a game between two players: the Learner, whose objective is to train a classifier, and the Adversary, whose objective is to modify some data in an attempt to evade detection by the classifier. In this section, we construct each player's objective function, before organising the players into a pessimistic bilevel program to model their interactions. The bilevel program is formulated by organising the players into a hierarchy, with one player taking the role of the leader and making their move first, before the other player, known as the follower, makes their move. As such, the follower might observe the leaders move before making their own, making the optimal move for the follower dependant on the leader's move. Since we consider tasks where an adversary seeks to evade detection by a classifier, we are assuming that classifier will have already been established before the adversary modifies their data. More importantly, we assume the adversary can observe the capabilities of the classifier when modifying their data. For this reason, it seems sensible to place the Adversary in the lower-level. From the learner's perspective, solving this game then equates to finding a classifier that is resilient to changes in the distribution.

Existing pessimistic bilevel approaches to adversarial learning allow the adversary to modify or generate data in an unrestricted nature \cite{benfield2024classificationstrategicadversarymanipulation}. This can make the adversary vulnerable to creating unrealistic data, which while optimal, might lose its intended meaning. While such a model is capable of creating the perfect adversary, it runs the risk of becoming overly pessimistic or data becoming incomprehensible. The learner's classifier, which is trained on this data, then suffers in performance. Other methods use regularisation terms to incentivise small movement by the adversary \cite{Brückner_Scheffer_2011}. However, their proposed solution method relies on strong assumptions about the convexity and uniqueness of the lower-level problem. Therefore, the choice of distance measure is restricted to strongly convex functions, such as the Euclidean norm. We address both of these points by constructing a pessimistic bilevel model with constraints in the lower-level that measure the similarity between the adversary's data and its initial value. We restrict this similarity to be above some pre-defined threshold. In this way, we have more control over the amount of freedom the adversary is allowed over their data. We make no assumptions about the convexity of the model or the uniqueness of the lower-level solution, giving us more freedom over the choice of similarity measure.


Let $D \in \mathbb{R}^{n q}, D=(D_1,\dots,D_n)^T$ be the static set of $n \in \mathbb{N}$ instances of data where each $D_i \in \mathbb{R}^q, \; i =1,\dots,n$, is a row vector containing the values of $q \in \mathbb{N}$ features, and let
$\gamma \in \{0,1\}^{n}$ be the corresponding collection of binary classes. The adversary possesses their own sample containing $m$ instances of the same $q$ features which they can modify in order to evade detection by the classifier. Let $X \in \mathbb{R}^{mq}$ be the data controlled by the adversary, defined as
\[X :=\begin{pmatrix}
    X_1^T \\
    \vdots \\
    X_m^T
\end{pmatrix},\]
where each $X_i, \; i = 1, \dots, m$, is a row vector of features. Let $Y \in \{0,1\}^m$ be the corresponding class labels of the adversary's data. The learner seeks to find the optimal weights, $w \in \mathbb{R}^q$, of some prediction function $\sigma : \mathbb{R}^q \times \mathbb{R}^q \rightarrow \mathcal{P}$, where $\sigma(w, x)$ gives the prediction of the label of a sample of data $x \in \mathbb{R}^q$. Here, $\mathcal{P} \subset \mathbb{R}$ is the prediction space. For classification tasks, for example, this could be $\mathcal{P} = (0,1)$ to represent probabilities, while for regression tasks, it could be the space $\mathcal{P} =\mathbb{R}$. Let $\mathcal{L} : \mathcal{P} \times \mathcal{P} \rightarrow \mathbb{R}$ be a loss function that penalises on incorrectly labelled data, we define the learner's objective function $F : \mathbb{R}^q \times \mathbb{R}^{m q} \rightarrow \mathbb{R}$ to be the sum of the loss on the static data and the adversarial data,
\begin{equation}
\label{eqn:UpperLevel}
    F(w,X) = \sum_{i=1}^{n} \mathcal{L}(\sigma(w, D_i), \gamma_i) + \sum_{i = 1}^m \mathcal{L}(\sigma(w, X_i), y_i).
\end{equation}

The adversary's objective is to identify the sample of data, $X$, which evades detection by the classifier. Existing pessimistic bilevel programs model the adversary with an unrestricted optimization problem where the objective function is opposite to the learner \cite{benfield2024classificationstrategicadversarymanipulation}. However, without restrictions on how the adversary can modify their data, the model is vulnerable to producing data that becomes nonsensical. While this might be the optimal data to evade detection, it is potentially unrealistic and not an accurate representation of the real-world adversarial data. As such, the classifier suffers in performance when implemented. To remedy this, we initialise our model by giving the adversary some real-word data which they can modify and define a set of constraints which restrict how much they can modify this data.

Let $\ell : \mathcal{P} \times \mathcal{P} \rightarrow \mathbb{R}$ be the adversary's loss function where, given some predictor weights $w$, $\ell(w, X)$ measures the success of a sample of data, $x \in \mathbb{R}^{q}$, at evading detection by the learner. We define the adversary's objective $f : \mathbb{R}^q \times \mathbb{R}^{m q} \rightarrow \mathbb{R}$ as the sum over some loss function applied to all instances of the adversary's data,
\begin{equation}
\label{eqn:lower-level obj}
    f(w, X) = \sum_{i=1}^n \ell(\sigma(w, X), y_i).
\end{equation}
Let $X^0 \in \mathbb{R}^{mq}, \; X^0 = (X^0_1, \dots,X^0_m)^T$ be the collection of the adversary's data before manipulation, where each $X^0_i \in \mathbb{R}^q, \; i=1,\dots,m$ is a row vector. To ensure that the adversary generates realistic data, we introduce the constraints $g : \mathbb{R}^{mq} \rightarrow \mathbb{R}^m$ defined as
\begin{equation}
    g(X) := \begin{pmatrix}
        g_1(X_1) \\
        \vdots \\
        g_m(X_m)
    \end{pmatrix},
\end{equation}
where each constraint function, $g_i : \mathbb{R}^q \rightarrow \mathbb{R}, \; i = 1,\dots,m$, is defined as
\begin{equation}
\label{eqn:constraints}
    g(X) := \delta - d(X_i, X^0_i),
\end{equation}
where $d : \mathbb{R}^{q} \times \mathbb{R}^{q} \rightarrow \mathbb{R}$ is some similarity function whose value decreases as the provided data diverge in similarity and $\delta$ is the desired similarity threshold.

By satisfying the constraints $g(X) \leq 0$, the adversary will identify some modified data that closely resembles its original position. This can prevent the adversary's data becoming nonsensical or losing its original sentiment. By initialising the adversary with some real-world data, this also ensures that the adversary's solution reflects the real-world. Identifying a suitable value of $\delta$ that ensures this will greatly depend greatly on the expected nature of the adversary. If we expect an aggressive adversary who is willing to make large changes, then we would select a low value of $\delta$, to allow the adversary a high amount of freedom. On the other hand, if we expect the adversary to make small perturbations to existing data, such as injecting strategic noise, then we would set $\delta$ to a large value that restricts the adversary to smaller movements. If we have access to a sample of adversarial data containing both of its versions, namely the original and the modified, then we could use these to estimate an appropriate value of $\delta$ by measuring the similarity between them. For example, with access to an untouched image and its compromised version that successfully evaded detection, we could set $\delta$ to the similarity score between the two, as we might expect other adversaries to follow a similar approach. In practice, however, this information is likely not available, for example, it is unlikely that we would possess all previous versions of a spam email. Instead, appropriate values of $\delta$ may be found through hyperparameter optimisation techniques such as grid search.

With an appropriate threshold identified, combining the objective function \eqref{eqn:lower-level obj} with these constraints, we simulate an adversary that finds data that evades detection by the classifier while remaining plausibly realistic. Collecting the two objectives together, we have the learner seeking to find the optimal weights of a predictor that accurately predicts the label of some data, while the adversary seeks to produce data that evades detection by the predictor and be assigned an incorrect label while under the constraint that their data must be similar to its original value. In the case that there are multiple optimal solutions to the adversary's problem, in other words, there are multiple instances of data that evade detection while satisfying the similarity score, then we take the pessimistic approach and assume that the adversary chooses the one which most harms the learner. This is to say that we assume that the adversary chooses the optimal solution that maximises the learner's objective function,
\begin{equation}
\label{eqn:bilevel}
    \min_{w \in \mathbb{R}^{q}} \max_{X \in S(w)} F(w,X),
\end{equation}
where $S(w)$ is the set of optimal solutions to the adversary's problem,
\begin{equation}
\label{eqn:lowerlevelset}
    S(w):= \argmin_{ X \in \mathbb{R}^{m q}} \left\{ f(w, X) \mid g(X) \leq 0, \;\; i = 1,\dots,m \right\}.
\end{equation}

The pessimistic bilevel optimisation program described by \eqref{eqn:bilevel}-\eqref{eqn:lowerlevelset} captures the antagonistic nature of the game between the learner and the adversary. With the learner in the upper-level, the solution to this game amounts to identifying the optimal weights of a predictor while considering how an adversary will react and modify their data to evade detection. The solution to this program then produces a predictor that accounts for adversarial manipulation during the training process, leading to more resilient predictors at time of implementation. By constraining the adversary to produce data that is similar to its original value, we prevent data becoming nonsensical or losing its original sentiment and so better reflects the movements of a real-world adversary.

For a practical illustration of the model \eqref{eqn:bilevel}-\eqref{eqn:lowerlevelset}, we apply it to text-based classification tasks. We present suitable choices of the loss functions and constraints before analysing some their properties to highlight the benefits of the novel model over the existing literature. In particular, we highlight the ability for our model to make use of the cosine similarity to measure adversarial movement, which is not achievable in current pessimistic approaches due to assumptions on convexity.

For a classification task with class labels $0$ and $1$, we design the prediction function to be the probability that a sample belongs to the positive class. The predictor space is hence set as $\mathcal{P} = (0,1)$. Let $x \in \mathbb{R}^q$ with $q$ be a sample of data containing $q$ features. For some weights $w \in \mathbb{R}^q$, we set the prediction function $\sigma$ as the sigmoid function,
\[\sigma(w, x) = \frac{1}{1+e^{-w^Tx}}.\]
Let $y \in \{0,1\}$ be the corresponding label of $x$. We set the learner's loss function, $\mathcal{L}$, to be the logistic regression loss function,
\begin{equation}
\label{eqn:loggistic_loss}
    \mathcal{L}(\sigma(w,x), y) := - y \log(\sigma(w,x)) - (1 - y) \log(1 - \sigma(w, x)).
\end{equation}
We design the adversary's loss function to be the logistic loss with opposite class labels,
\begin{equation}
\label{eqn:adversarys_loss}
    \ell(\sigma(w,x), y) := (y - 1) \log(\sigma(w,x)) - y \log(1 - \sigma(w, x)).
\end{equation}

When embedding text-based data, a common strategy is to use cosine similarity to identify appropriate vector representations while considering the similarities between words. See, for example, Google's BERT \cite{BERT} which embeds the text in such a way that the angle between words with a similar definition will be small and hence receive a high cosine similarity. In this way, we can allow the adversary to change some words in the email without losing the intended message. Consider, for example, the legitimate and spam email in Table \ref{tbl:email_examples}.
\begin{table}[H]
\centering
\begin{tabular}{c | c} 
Email message & Spam indicator\\
 \hline
Thanks for your order! You can click here to track it & No \\
Your account has been compromised! Click here to recover it & Yes
\end{tabular}
\caption{Example emails}
\label{tbl:email_examples}
\end{table}

We embed these emails in the vector space $\mathbb{R}^{128}$ using Google's BERT \cite{BERT} and train the logistic regression classifier in \eqref{eqn:loggistic_loss} to successfully distinguish between the legitimate and spam. Now consider how an adversary might update the spam email to evade this classifier. For example, by simply changing the phrasing of the email to the following
\[\text{``Your account is in danger! Click here to fix it."},\]
we now evade detection. Clearly, this new email is very similar to the original. We have replaced the phrasing with words that share a similar definition and so retain the intended message. This is reflected in a high cosine similarity score of 0.992 between this new email and the original. Note that under the adversary's loss function, given in \eqref{eqn:adversarys_loss}, this modified email scores 0.869. Now consider changing the email to a string of nonsensical letters, such as
\[\text{``aaaaa."}.\]
This message contains no meaningful message and hence will not benefit the adversary. This is supported by the lower cosine similarity score of 0.735. However, it scores a loss of 0.443, which is considerably less than the meaningful transformation. Therefore, under an unrestricted model, the adversary would favour the nonsensical message over the meaningful one. Clearly, a real adversary is more likely to make changes that retain the original message. To ensure we reflect this in our model, we can constrain the adversary to finding a message that scores a cosine similarity above, for example, 0.9. In this way, we prevent the adversary from constructing an email that is too dissimilar from its original.

To incorporate the cosine similarity into the model, we can set the similarity function, $d$ as the cosine similarity between the adversary's data and its original value. Let $\theta_{x, x^0}$ be then be the angle between $x$ and its original value $x^0$. We set the similarity function $d : \mathbb{R}^{q} \times \mathbb{R}^q \rightarrow (-1,1)$ as
\[d(x,x^0) = \cos \left ( \theta_{x, x^0} \right ) = \frac{x \cdot x^0}{\Vert x \Vert \Vert x^0 \Vert}.\]
The cosine similarity is defined in the range $(-1,1)$, where $d(x,x^0) = 1$ indicates identical similarity and $d(x,x^0) = -1$ indicates opposite similarity. The constraint functions in \eqref{eqn:constraints} then become,
\begin{equation}
\label{eqn:cosine}
    g_i \left ( X_i \right ) = \delta - d(X_i, X^0_i) = \delta - \frac{X_i \cdot X_i^0}{\Vert X_i \Vert \Vert X_i^0 \Vert}, \; i \in \{1,\dots,m\}.
\end{equation}
where $\delta \in (-1,1)$ is the desired similarity threshold. The constraints $g(X) \leq 0$ then ensure that the cosine similarity between the adversary's data, $X$, and its original value, $X^0$, is greater than $\delta$. Note that the cosine similarity, as well as its first and second derivatives are undefined when either $X^0_i = 0$ or $X_i = 0$ for some $i \in \{1,\dots,m\}$. However, we can reasonably discount any issues that could arise from this since both cases correspond to empty data.

We now explore some of the properties of the model described by \eqref{eqn:bilevel}-\eqref{eqn:lowerlevelset} for text-based classification tasks that use cosine similarity. Through a simple example in Proposition \ref{prop:unique_sols} below, we demonstrate that the lower-level problem possesses multiple optimal solutions. Note that this proposition applies to any case where the prediction function, $\sigma$, can be expressed as a function $\Omega : \mathbb{R} \rightarrow \mathcal{P}$ of the linear combination $w_0 + w^Tx$. For the case of the sigmoid function, as in this application, this holds when $\Omega(v) = \frac{1}{1+e^{-v}}$, for example.

\begin{proposition}
\label{prop:unique_sols}
    Let $w_0 \in \mathbb{R}, w \in \mathbb{R}^q, w_i \neq 0 \, \forall \, i \in \{1,\dots,q\}$ where $q > 1$ and let $X \in \mathbb{R}^{m q}, X^0 \in \mathbb{R}^{m q}$ with corresponding labels $Y \in \{0,1\}^m$. Let $f(w, X)$ be defined as in \eqref{eqn:lower-level obj} where $\sigma$ is some prediction function that can be expressed as a function $\Omega : \mathbb{R} \rightarrow \mathcal{P}$ of the linear combination $w_0 + w^Tx$,
    \[\sigma(w,x) = \Omega(w_0 + w^Tx),\]
    where $x \in \mathbb{R}^q$ and $\mathcal{P} \subset \mathbb{R}$. Then the lower-level problem defined by \eqref{eqn:lowerlevelset} admits multiple optimal solutions.
\end{proposition}

The existing pessimistic bilevel program with restriction on adversarial movement as proposed by \cite{Brückner_Scheffer_2011} is shown to possess a unique solution to the lower-level problem which plays a vital role in the proposed solution method. As such, it is not possible to incorporate cosine similarity constraints in the existing pessimistic model.

Secondly, we demonstrate through Proposition \ref{prop:concavity} below that feasible region of the lower-level problem, defined by the constraints, is non-convex in $X$.

\begin{proposition}
\label{prop:concavity}
    Let $\delta \in (-1, 1)$ and let $X^0 \in \mathbb{R}^{m q}$ be such that for any $i \in \{1,\dots,m\}$ we have $X^0_{ij} = X^0_{ik} \; \forall \; j,k \in \{1,\dots,q\}$. Let $g$ be defined as in \eqref{eqn:constraints}, where the similarity function $d$ is defined to be the cosine similarity as defined in \eqref{eqn:cosine}, then the set defined by $\{X : g(X) \leq 0\}$ is non-convex.
\end{proposition}

\begin{figure}[H]
    \centering
    \includegraphics[scale = 0.3]{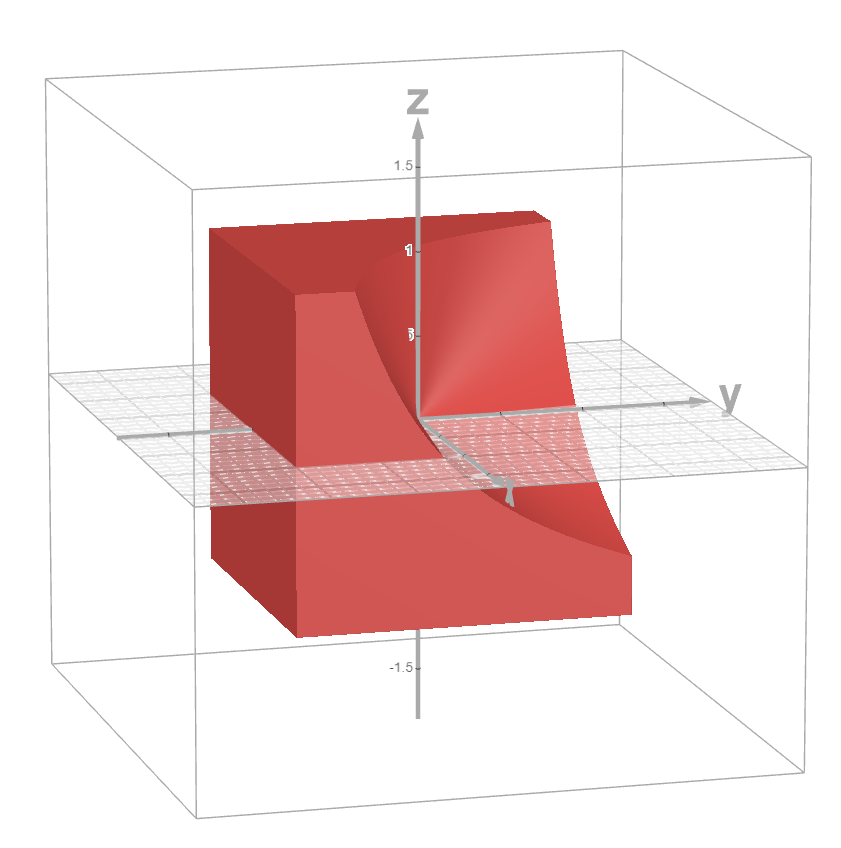}
    \caption{Illustration of the non-convexity of the lower-level feasible region $\{X \in \mathbb{R}^{3} : g(X) \leq 0\}$ where $X^0 = (-1,-1,-1)^T$ and $\delta = -0.8$.}
    \label{fig:concavity}
\end{figure}

We illustrate Proposition \ref{prop:concavity} by plotting the feasible region defined by the lower-level constraints of a simple three-dimensional example for a single samples of data where $X^0 = (-1,-1,-1)^T$ and $\delta = -0.5$ in Figure \ref{fig:concavity}. It is clear from the plot that the feasible region lower-level problem is non-convex.

To summarise, throughout this section we introduce the pessimistic bilevel model described by \eqref{eqn:bilevel} - \eqref{eqn:lowerlevelset} and demonstrate how it can be applied to text-based adversarial classification tasks. By substituting the predictor function for the sigmoid function and the upper-level loss function for the logistic loss function, we constructed a formulation that allows the learner to train a classifier. In the lower-level, we also substituted the loss function for the logistic loss, except with opposite class labels so that the adversary optimises their data towards the opposite class. The rational here being that the adversary is attempting to have their data misclassified as the opposite label. For example, a spammer wishes to have their spam email as a legitimate email. Finally, we demonstrated that for text-based classification, the similarity function can be substituted for the cosine similarity. Since text-based data, such as emails, can be embedded using the cosine similarity as a measure between words with similar dictionary definitions, such constraints can be used to ensure that the adversary manipulates their data such that the original message is not lost.

We then demonstrated through Propositions \ref{prop:unique_sols} and \ref{prop:concavity} that, under these substitutions, the lower-level problem described by \eqref{eqn:lowerlevelset} is non-convex and can possess multiple optimal solutions. Such a pessimistic bilevel model has not yet been proposed since existing pessimistic models rely on uniqueness of lower-level solutions and convexity to find solutions to the program.


\section{Solving the problem}
\label{sctn:solution method}
In section \ref{sctn:bilevelModel}, we first constructed a general form of a pessimistic bilevel program with lower-level constraints to model adversarial evasion scenarios under restricted data manipulation. Subsequently, we illustrated this model on a text-based classification task  and how the introduction of lower-level constraints allows us to better control how the adversary manipulates their data, allowing for more meaningful transformations that better reflect reality. Through some motivating examples, we show in Propositions \ref{prop:unique_sols} and \ref{prop:concavity} that the lower-level problem in our pessimistic bilevel program  \eqref{eqn:bilevel} -- \eqref{eqn:lowerlevelset} does not necessarily possess a unique solution and can be non-convex, respectively. Hence, the use of the traditional Karush-Kuhn-Tucker reformulation (see, e.g., \cite{dempe2014necessary,benchouk2025scholtes}) to solve problem \eqref{eqn:bilevel} -- \eqref{eqn:lowerlevelset} is generally impossible. This makes solving the problem even harder. Moreover, in this context, any algorithm to solve the problem is very likely to only compute stationary points, considering the fact that it is non-convex. Hence, while accounting for the observations detected in Propositions \ref{prop:unique_sols} and \ref{prop:concavity}, a key aim of this section is to construct a system of equations that characterizes the stationary points of problem \eqref{eqn:bilevel} -- \eqref{eqn:lowerlevelset}, which will then be embedded in a general algorithmic framework for the training process of a text-based adversarial learning scenario, to illustrate the practical use of the general mathematical model from Section \ref{sctn:bilevelModel}.



To proceed here, note that a point $\bar w$ will be said to be a local optimal solution for problem \eqref{eqn:bilevel}--\eqref{eqn:lowerlevelset} if there is a neighborhood $W$ of the point such that 
\begin{equation}\label{eq:OptimalConcept}
    \varphi_p(\bar w) \leq \varphi_p(w) \;\, \mbox{ for all }\; w\in W,
\end{equation}
where the $\varphi_p$ denotes the following \textit{two-level value function} (concept first introduced and studied in \cite{dempe2012sensitivity}):
\[
\varphi_{\textup{p}}(w):= \max\{F(w,X)\,|\,X\in S(w)\}.
\]


From now on, we assume that the adversary's original data $X^0 \in \mathbb{R}^{m q}$ is such that $X^0_i\neq 0$ for all $i=1, \ldots, m$, as otherwise, the distance function defining $g$ would be undefined. We can safely make this assumption as setting $X^0$ to some data value such that $X^0_i = 0$ for some $i \in \{1,\dots,m\}$ would correspond to an instance of empty data. Since adversary uses $X^0$ as a comparison to help ensure they are producing realistic and meaningful solutions, we can remove any empty data prior to experiments as they would not provide any meaningful comparison.


Now, if $w$ is a local optimal solution of problem \eqref{eqn:bilevel}--\eqref{eqn:lowerlevelset}, in the sense of \eqref{eq:OptimalConcept}, and we assume that a suitable set of qualification conditions are satisfied (see the full detail  in Theorem \ref{thm:system} and the corresponding proof in Appendix \ref{Proofs}), then there exists a Lagrange multiplier vector $(\lambda, \beta, \hat \beta)$ such that the following stationarity conditions are satisfied:
\begin{subequations}\label{eq:KKT_pbpp_abstract_rerefined}
	\begin{align}
		\label{KTpes6}
        \nabla_w F(w,X) & = 0,\\
        \label{KTpes6.1}
		\nabla_X F(w, X) - \lambda \nabla_X f(w, X) - \nabla g(X)^\top \beta & = 0,\\
		\label{KTpes7}
		\nabla_X f(w, X) + \nabla g(X)^\top\hat\beta& = 0,\\
		\label{KTpes8}
		\hat\beta \geq 0, \quad g(X)\leq 0,\quad \hat\beta^\top g(X)&=0,\\
        \label{KTpes8.1}
      \lambda \geq 0,\;\,  \beta\geq 0,\quad g(X)\leq 0,\quad \beta^\top g(X)&=0,
 	\end{align}
\end{subequations}
for some of the adversary's data matrix $X\in  S_p(w)$ such that $X_i\neq 0$ for all $i=1, \ldots, m$ with
\begin{equation}\label{eq:Sp-Incl}
    S_p(w) := \argmax_{X \in \mathbb{R}^{m q}} \{F(w, X) \, | \, X \in S(w) \}.
\end{equation}
Following the proof of Theorem \ref{thm:system}, it should be easy to see that for a local optimal solution $w$ of problem \eqref{eqn:bilevel}--\eqref{eqn:lowerlevelset}, under suitable assumptions (that the technicality of is out of the main scope of this paper), for some $X\in  S_p(w)$ (to be obtained by some technical assumption called inner semicontinuity (see Appendix \ref{Proofs}, it is necessary that the conditions \eqref{KTpes6.1}--\eqref{KTpes7} and \eqref{KTpes8.1} are satisfied. Hence, if one looks carefully, it would become apparent that inclusion \eqref{eq:Sp-Incl} is already characterised in the system \eqref{KTpes6}--\eqref{KTpes8.1}; see some relevant discussion in \cite[Section 5]{benchouk2025scholtes}.

Similarly to the adversary's original data, it is necessary that we have $X_i\neq 0$ for all $i=1, \ldots, m$, as otherwise, the lower-level constraint function $g$ would also be undefined. For a given set of the training parameters, $w$, the set $S(w)$ contains the set of optimal data matrices that evade detection by the learner. The set $S_p(w)$ contains the subset of these solutions that not only evade detection, but also do the most damage to the learner's classifier. More specifically, these solutions result in the highest value of the learner's objective function. It is reasonable to assume that under any reasonably performing classifier, the adversarial instance which deals the most harm to the learner would be non-zero since empty data would easily be detected as suspicious. Further to this, under an appropriate choice of the similarity threshold, $\delta$, we can expect that an empty instance of data would not provide a sufficiently high similarity score since empty data should be considered meaningless.


To solve problem \eqref{eqn:bilevel}--\eqref{eqn:lowerlevelset}, we will solve the system \eqref{KTpes8}--\eqref{KTpes8.1}. Hence, to facilitate the process of solving this system, we will transform the complementarity conditions in \eqref{KTpes8}--\eqref{KTpes8.1} into a system of equation. 
For a compact notation, let us first introduce block variables by means of
\[
        z:= \begin{bmatrix}
            w \\
            X
        \end{bmatrix} \in \R^{q+mq},\qquad
	\xi:=\begin{bmatrix}\beta\\\hat\beta\\\lambda\end{bmatrix}\in\R^{2b+1},
\]
and Lagrangian-type functions $L^\textup{p}_w,L^\textup{p}_X,\ell^\textup{p}\colon(\R^q \times \R^{m q})\times\R^{2b+1}\to\R$ as stated below:
\begin{align*}
	\forall z\in\R^{q + m q},\,\forall\xi\in\R^{p+2q+1}\colon\quad
	L^\textup{p}_w(z,\xi)   &:= F(w,X),\\ 
	L^\textup{p}_X(z,\xi) 	&:= F(w,X)-\lambda f(w,X)-\beta^\top g(X),\\
	\ell^\textup{p}(z,\xi) 	&:= f(w,X)+\hat\beta^\top g(X).
\end{align*}
Now, we can set
\[
	\calG(z)
	:=
	\begin{bmatrix}
		g(X)\\g(X)\\0
	\end{bmatrix},
	\qquad
	\calH(z,\xi)
	:=
	\begin{bmatrix}
		\nabla_w L^\textup{p}_w(z,\xi)\\
		\nabla_X L^\textup{p}_X(z,\xi)\\
		\nabla_X \ell^\textup{p}(z,\xi)
	\end{bmatrix}.
\]

Based on these notations, the system \eqref{KTpes6}--\eqref{KTpes8.1} can be equivalently written as follows (with $(m+1)q+2b+2$ variables and $(2m+1)q+2b+2$ equations): 
\begin{equation}
\label{eqn:inequalities}
\calH(z, \zeta) = 0, \;\;    \zeta \geq 0, \, \calG(z) \leq 0, \, \zeta^T \calG(z) = 0.
\end{equation}
To fully convert this system, which combines a nonlinear equation and complementarity conditions, into a system of equations, we introduce the well-known a the  Fischer-Burmeister function \cite{fischer1992special}, defined as
\begin{equation}
\label{eqn:FB}
    \vartheta_{\text{FB}} (a,b) = \sqrt{a^2 + b^2} - a - b \;\mbox{ for }\; (a,b)\in \mathbb{R}^2.
\end{equation}
The transformation process to be done here is based on  the following important property:
\[\vartheta_{\textup{FB}}(a, b) = 0 \iff a, b \geq 0, \, a^Tb = 0.\]
Therefore, we can encode \eqref{eqn:inequalities} into a system of equalities by
\begin{equation}
\label{eqn:system}
    \Phi(z, \zeta) = \begin{pmatrix}
        \calH(z, \zeta) \\
        \vartheta_{\textup{FB}} \left ( \zeta, -\calG(z) \right )
    \end{pmatrix} = 0,
\end{equation}
where the use of the Fischer-Burmeister here should be understood vectorwise; i.e., 
\[
\vartheta_{\textup{FB}} \left (\zeta, -\calG(z) \right ):=\left(\begin{array}{c}
     \vartheta_{\textup{FB}} \left ( \zeta_1, -\calG_1(z) \right )\\
     \vdots\\
     \vartheta_{\textup{FB}} \left ( \zeta_T, -\calG_T(z) \right )
\end{array} \right) \;\mbox{ with }\; T:= 2b+1.
\]

{\color{green}

}

Given that the system \eqref{eqn:system} is overdetermined (precisely, it has $m$ more equations than variables),  the Levenberg-Marquardt method is a suitable scheme compute its zeros. Specifically, we make use of the global nonsmooth Levenberg–Marquardt method for mixed nonlinear complementarity systems developed in \cite{LevenbergMarquardt}, as its theoretical convergence is proven with a suitable generalized derivative concepts for the Fischer-Burmeister function; namely, the Newton-derivative (represented by $D_N$) is used to extend differentiability to the function $\Phi$ defining the  system \eqref{eqn:system}:
\[
D_N \Phi(z, \zeta) := \begin{pmatrix}
    \nabla \mathcal{H} (z, \zeta) \\
    D_N \vartheta_{\text{FB}} (\zeta, -\mathcal{G}(z))
\end{pmatrix},
\]
where
\[
D_N \vartheta (\zeta, -\mathcal{G}(z)) := \begin{cases}
    \left( \frac{\sqrt{2}}{2} - 1, \frac{\sqrt{2}}{2} - 1 \right ) & \mbox{ if } (\zeta,-\mathcal{G}(z)) = (0,0), \\
    \left ( \frac{\zeta}{\sqrt{\zeta^2 + (\mathcal{G}(z))^2}} - 1, \; -G^\prime(z) \left [ \frac{1}{\sqrt{\zeta^2 + (\mathcal{G}(z))^2}} - 1 \right ] \right ) & \text{otherwise}.
\end{cases}
\]
At iteration $k$, the core step of the algorithm is to find $d^k$ that solves the linear system of equations
\[
(D_N \Phi(z^k)^T D_N \Phi(z^k) + v_k \mathbb{I})d = -D_N \Phi(z^k)^T \Phi(z^k),
\]
where $\mathbb{I}$ denotes the identity matrix of suitable dimension and $v_k$ is a positive parameter that could be selected in different ways. 
As demonstrated through extensive convergence analysis of the Levenberg--Marquardt algorithm in \cite{stopping_criteria_1, stopping_criteria_2}, it is often beneficial to introduce a stopping criteria that halts the algorithm when it stagnates and shows little change in the value of the objective. Therefore, in algorithm \ref{alg:LevenbergMarquardt}, we present a modified version with an additional stopping criterion. In particular, we introduce steps 14-16 which halt the algorithm when the ratio between the value of the objective at the current iteration and its value in the previous iteration exceeds some $\eta \in (0,1)$. This stopping criteria is only enacted if the we exceed some set number of iterations $K \in \mathbb{N}$. 

To measure progress of the algorithm towards a solution for \eqref{eqn:system},  the merit function 
\[
\Psi_{\textup{FB}}(z, \zeta) := \frac{1}{2} ||\Phi(z, \zeta)||^2
\]
is used to compute a step size at each iteration to promote the global convergence. 
On important nice thing about the function $\Psi_{\textup{FB}}$, that facilitates the latter point (computing the step size) is that it is continuously differentiable on open neighbourhoods of given points from its domain. Then considering the Newton derivative formula above, one can check that 
\[
\nabla \Psi_{\textup{FB}}(z, \zeta) = D_N\Psi_{\textup{FB}}(z, \zeta)^\top \Psi_{\textup{FB}}(z, \zeta).
\]
A precise description of the Levenberg--Marquardt algorithm briefly described here to solve the system \eqref{eqn:system} is given in Algorithm \ref{alg:LevenbergMarquardt}. For a detailed analysis and the convergence theory, see \cite{LevenbergMarquardt}. 

We now introduce, in Algorithm \ref{alg:Framework}, a general framework for the training process of a text-based adversarial learning scenario. This framework is intended for use with time-stamped data, such as emails which record the day and time of receipt. We assume that data is organised chronologically with the expectation that the nature of some the data changes in some adversarial way over time. For example, we would expect spam emails to evolve over time as the adversary updates their strategies to overcome evolving email filters.

\begin{algorithm}[H]
\caption{Framework to construct and solve text-based adversarial learning scenarios}\label{alg:Framework}
\begin{algorithmic}[1]
\Require Text-based dataset $\mathcal{X}$
\State Embed the text-based data via the cosine similarity
\State Order the data chronologically
\State Divide the data, $\mathcal{X}$, into a training set, $\mathcal{X}_\text{train}$ and a testing set, $\mathcal{X}_\text{test}$.
\State Further divide the test data, by time, into $K$ subsets, $\mathcal{X}_{\text{test}}^1, \dots, \mathcal{X}_{\text{test}}^K$.
\State Divide the training data into two distinct subsets: a set of static training data, $D \subset \mathcal{X}_\text{train}$, and a set of starting points of the adversary, $X^0 \subset \mathcal{X}_\text{train}$, where $D \cup X^0 = \mathcal{X}_\text{train}$ and $D \cap X^0 = \emptyset$.
\State Randomly generate the start point $w^0 \in \mathbb{R}^p$ for the learner.
\State Construct the system $\Phi(z,\zeta)$ as in \eqref{eqn:system}.
\State Solve the system by the Levenberg-Marquardt algorithm \ref{alg:LevenbergMarquardt} and extract the optimal weights of the learner's classifier $w^*$.
\State \textbf{return} $w^*$
\end{algorithmic}
\end{algorithm}

With the data organised chronologically, we take the earliest occurring instances to form the training set. Later instances then form the test set as we would expect theses instances to have been subjected to adversarial influence. Unlike typical approaches to classifier training, it is perhaps more appropriate here to reserve the larger portion of the data for testing. In this way, we can divide the test data in multiple distinct sets and assess the quality of the classifier over time, as detailed in step 4. A portion of the training data is assigned to the adversary in the lower-level. This forms the initial value of the data that they can manipulate. The system of equations in \eqref{eqn:system} is formulated and solved with Algorithm \ref{alg:LevenbergMarquardt} before extracting the optimal weights to construct a classifier that is resilient to adversarial manipulation.

\section{Numerical experiments}
\label{sctn:Experiments}
In this section, we assess the resilience of the classifier resulting from the pessimistic bilevel program. We design two experiments that measure the classifier's performance on unseen text-based data from the future. The experiments are designed on chronologically ordered data that we expect to naturally evolve over time as adversaries adapt to overcome evolving classifiers. We make use of two datasets, the first, \textit{TREC}, is an email corpora provided for the 2006 NIST Text Retrieval Conference, \cite{trec06}. This is a collection of spam and legitimate emails received throughout the years of 1993--2006, which, when compiled in a dataset, contain a total of 68338 emails. The second dataset, \textit{Amazon}, is a collection of 3185845 fake and legitimate reviews for cell phones and their accessories made between the years of 1998--2014, collected by \cite{amazon}. We organise both data sets chronologically and take the earliest 2000 samples as the training set. The remaining data are separated by year to form multiple distinct test sets. The data are converted into vectors in the space $\mathbb{R}^{512}$ using Google's BERT \cite{BERT}. Note that due to computational limitations, we restrict our attention in the \textit{Amazon} data to the time range of $1998 - 2012$.

Adversarial datasets are often imbalanced, and more importantly, it is possible for this imbalance to change over time. For example, when the \textit{TREC} data is divided, by year, into 8 distinct test sets, we see the proportion of spam emails ranging between $20.7\%$ and $39.4\%$. While the $F1$ score is a popular choice for performance measurement, its asymmetry towards the positive class can lead to misleading results. When a dataset has a positive class majority, a classifier that accurately detects the positive class (e.g. spam emails) will record a high F1 score even if its ability to accurately detect the negative class (e.g. legitimate emails) is poor, making it difficult to identify which classifier performs best. For example, an email filter might play it safe by simply classifying every email as spam. On a spam-majority test set, this classifier will record a high F1 score while in reality providing no meaningful classifications. This makes comparisons across unevenly proportioned test sets unfair. Therefore, we instead measure a classifier's performance by the symmetric $P_4$ metric \cite{p4_metric}, which is defined by
\[P_4 := \frac{4 \cdot \text{TP} \cdot \text{TN}}{4 \cdot \text{TP} \cdot \text{TN} + (\text{TP} + \text{TN}) \cdot (\text{FP} + \text{FN})},\]
where TP, TN, FP and FN represent the counts of true positives, true negatives, false positives and false negatives, respectively. In this way, we gain a better understanding of a classifier's ability to discern between classes.

\begin{figure}[H]
    \centering
    \includegraphics[scale = 0.37]{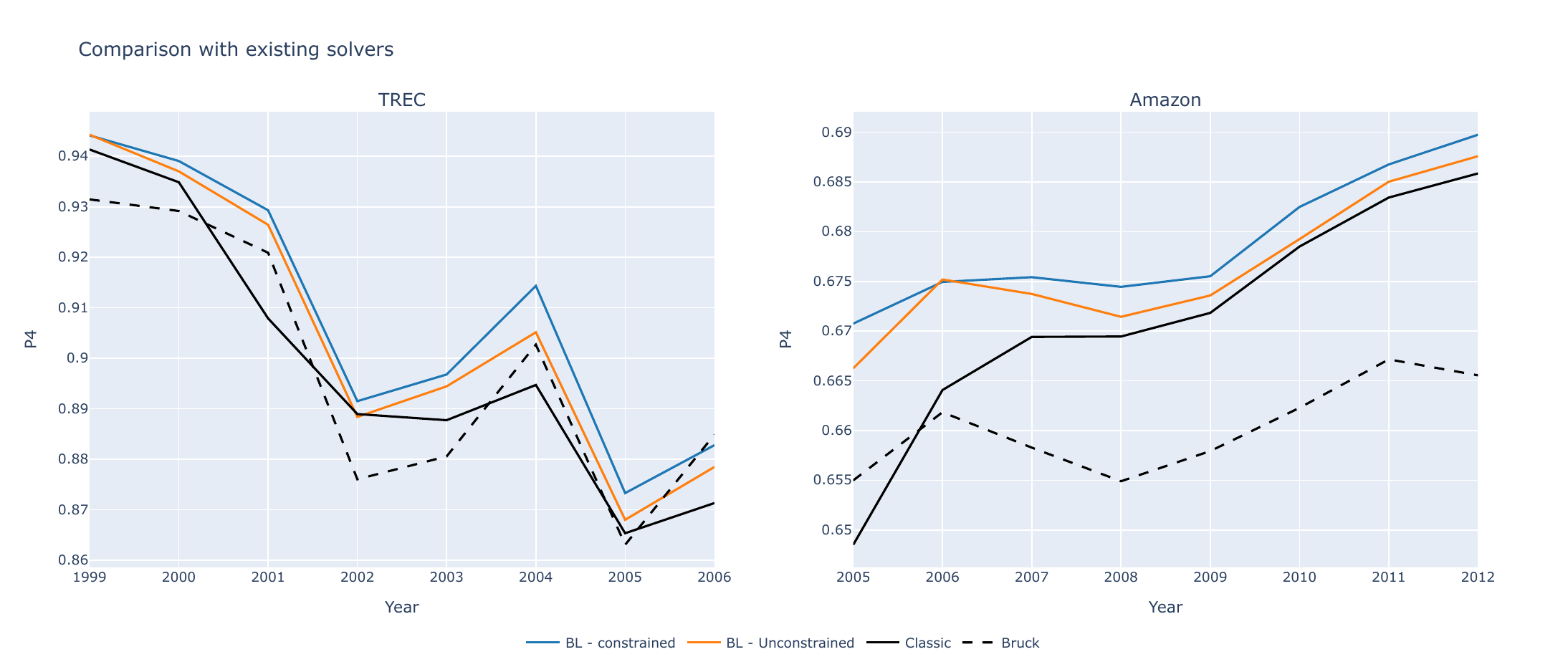}
    \caption{Comparison to classic model}
    \label{fig:comp_to_classic}
\end{figure}

To properly assess the our adversarially-trained classifier, we compare it to a classically-trained logistic regression classifier which does not consider any potential adversarial manipulations during its training process. The performance of this classifier serves as a baseline comparison. The $P_4$ performance of the classic classifier, named \textit{classic}, on the various test sets is displayed in Figure \ref{fig:comp_to_classic}. Recall that, relative to the training set, the test sets contain instances from the future. On the spam email dataset, \text{TREC}, we see a general trend of decreasing performance over time as the adversary modifies their approach, as expected. The same is not true, however, for the \textit{Amazon} dataset, perhaps suggesting less strategic change by adversaries over time.

The weights of the \textit{classic} classifier are taken as the starting point, $w^0$, of the bilevel model. We then construct and solve system \eqref{eqn:system}, grid searching various sizes of the adversary's sample: $m = \{1,2,5,10\}$ and the similarity threshold: $\delta = \{0.9, 0.99, 0.999\}$, to find the constrained bilevel classifier, named \textit{BL - Constrained}. The best classifiers for each year are given in Figure \ref{fig:comp_to_classic}. Additionally, we plot the performance of the unconstrained bilevel model from \cite{benfield2024classificationstrategicadversarymanipulation}. On both datasets, we see the bilevel models out-perform the classic model across all test sets. We also compare to the pessimistic bilevel model proposed in \cite{Brückner_Scheffer_2011}, named \textit{Bruck}. Despite beating the \textit{classic} classifier on numerous test sets in the \textit{TREC} experiment, we see a general trend of the \textit{Bruck} classifier performing worse than the bilevel models, except for the year 2006, suggesting that the bilevel models better capture the adversarial nature of the spam emails. On the \textit{Amazon} test sets, we see the \textit{Bruck} model struggle to perform better than the classic model, perhaps because the data are imbedded with the cosine similarity, while the \textit{Bruck} model is restricted to measuring adversarial movement by the Euclidean norm. The bilevel models, on the other hand, manage to capture this movement, with the constrained model consistently performing best, supporting the use of the cosine similarity in the constraints to measure adversarial movement.

We recall that the purpose of the constraints is to ensure that the data generated by the adversary are sufficiently similar to some real-world data. The numerical experiments in \cite{benfield2024classificationstrategicadversarymanipulation} demonstrated that many start points were required in order to identify a classifier that performed well under the unconstrained bilevel model. Therefore, to better understand the behaviour of the constrained bilevel model and for a deeper comparison between it and the unconstrained bilevel model, we investigate various values of its hyperparameters, namely the number of instances that the adversary can modify, given by $m$ and the similarity threshold, $\delta$. We then also compare their performance across many start points.

\begin{figure}[H]
    \centering
    \includegraphics[scale = 0.38]{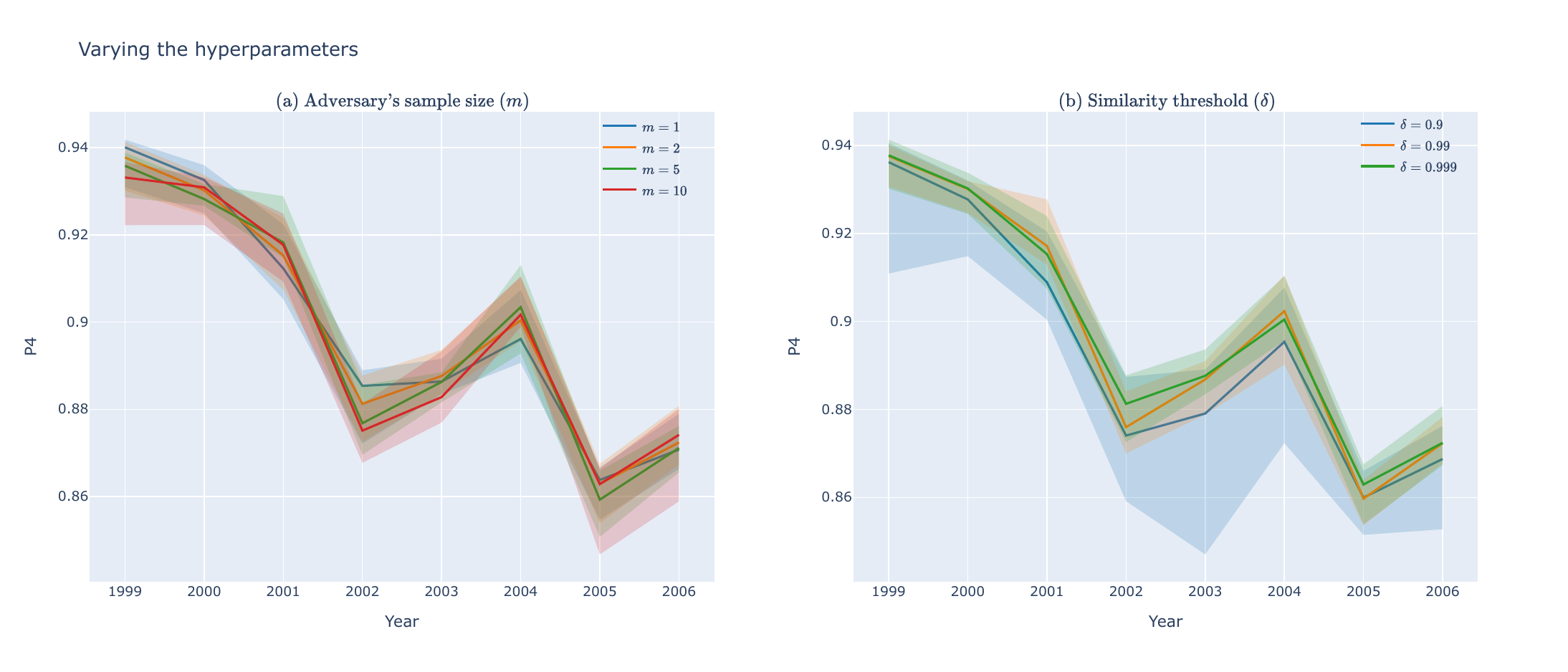}
    \caption{Investigation in to the affect of the similarity threshold $(\delta)$}
    \label{fig:varying hyperparameters}
\end{figure}

Beginning with the adversary's sample size, $m$, Figure \ref{fig:varying hyperparameters} (a) displays the effect of varying $m$ on the performance of the classifier. For each value of $m$, 100 starting points are tested and the median performance, as well as the $5\%$ and $95\%$ percentile, are plotted. For both values of $m$, the threshold is fixed to $\delta = 0.999$. The percentiles suggest a fairly similar spread in variation about the median. However, when comparing the median values, the results begin to suggest that a higher value of $m$ can lead to greater variation from the baseline classifier. To measure this variation, we calculate the distance of the $P_4$ performance of the constrained bilevel model to the baseline \textit{classic} model. This distance is measured for each value of $m$ with the threshold fixed to $\delta = 0.999$ and recorded in Table \ref{tbl:deviation}(A). We find a clear pattern of increasing variation as $m$ increases. These results, combined with the performance plots in Figure \ref{fig:varying hyperparameters} (a), suggest that giving the adversary a larger sample size leads to greater influence of the training process. This has the potential to yield higher performance, as demonstrated in years 2001 and and 2006 where $m=10$ performed best. However, there are also cases where the adversary has been given too much influence, leading to a drop in performance such as in years 2002 and 2003 where $m = 10$ performed worse. These classifiers are perhaps too pessimistic due a large adversarial influence during training. It is clear that a balance needs to be struck between enough adversarial influence that we account for natural changes in the data, but not too much that we become overly pessimistic and overestimate the abilities of adversaries.

\begin{table}[!htb]
    \begin{subtable}{.5\linewidth}
      \centering
        \begin{tabular}{l | c} 
         m & Variation from baseline \\
         \hline
        1   & 0.0068 \\
        2   & 0.0147 \\
        5   & 0.0211 \\
        10   & 0.0214 \\
        \end{tabular}
        \caption{Adversary's sample size $(m)$}
    \end{subtable}%
    \begin{subtable}{.5\linewidth}
      \centering
        \begin{tabular}{l | c} 
         $\delta$ & Variation from baseline \\
         \hline
        0.9   & 0.0203 \\
        0.99   & 0.0196 \\
        0.999   & 0.0137 \\
        \end{tabular}
        \caption{Similarity threshold $(\delta)$}
    \end{subtable}
    \caption{Deviation from baseline model (euclidean norm)}
    \label{tbl:deviation}
\end{table}

A similar pattern is found when varying the similarity threshold, $\delta$. With the adversary's sample size fixed to $m=2$, we again test 100 randomly selected start points for each value of $\delta \in \{0.9, 0.99, 0.999\}$. Figure \ref{fig:varying hyperparameters} (b) displays the median performance as well as the $5\%$ and $95\%$ percentile. Table \ref{tbl:deviation} (B) displays the variation of the median from the baseline model for each value of $\delta$. It is clear from Table \ref{tbl:deviation} (B) that as we increase the similarity threshold, we observe lower variation from the baseline. This is supported by the larger inter-percentile range for $\delta = 0.9$ than for the larger values of $\delta$, as shown in Figure \ref{fig:varying hyperparameters} (b). These results also demonstrate how giving the adversary too much freedom leads to lower performance. The median performance of the classifier where $\delta = 0.9$ consistently performs worst, supporting the idea that the adversary's data has become too far detached from reality and does not sufficiently represent the movements of a real adversary. We again conclude that a balance needs to be struck between allowing the adversary enough freedom that they can manipulate their data to evade detection, while not too much freedom that their data become unrealistic.

\begin{figure}[H]
    \centering
    \includegraphics[scale = 0.7]{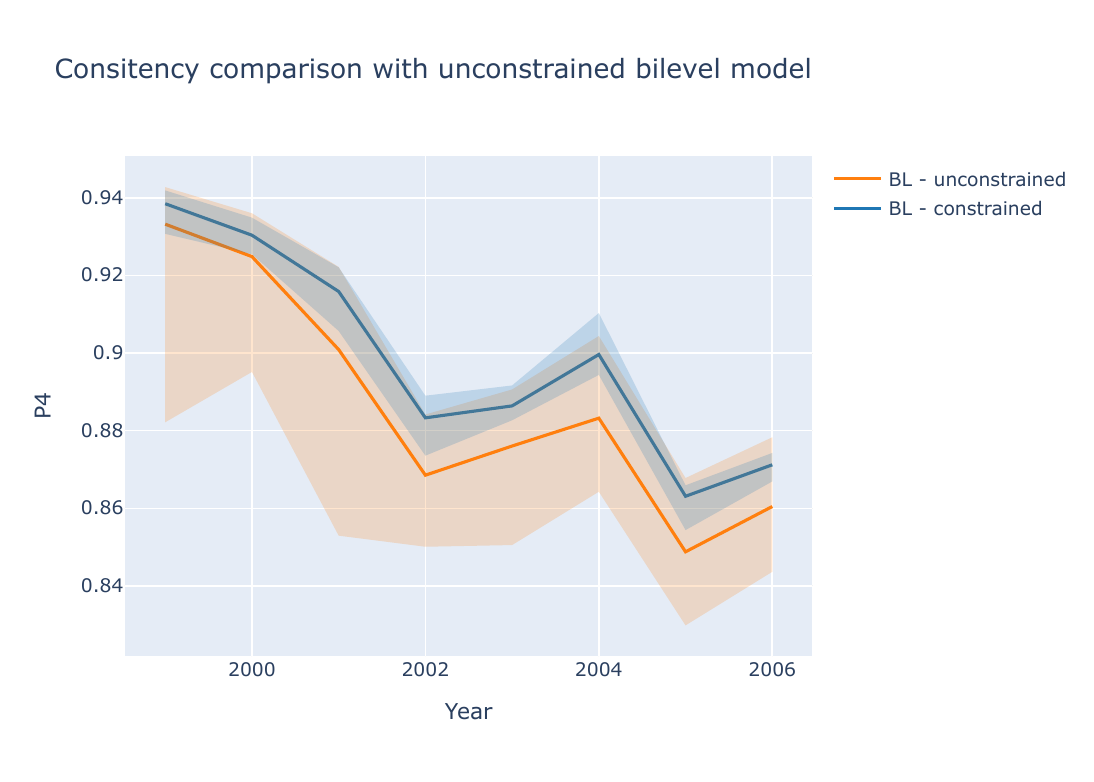}
    \caption{Comparison to unconstrained bilevel model for many starting points}
    \label{fig:comp_to_first_paper}
\end{figure}

Finally, we compare the constrained bilevel model to the unconstrained bilevel model across multiple starting points. Figure \ref{fig:comp_to_first_paper} displays the median performance for both bilevel models as well as the $5\%$ and $95\%$ percentiles across 100 random starting points. For the constrained model we choose the adversary's sample size to be $m=2$ and the similarity threshold to be $\delta = 0.999$. It is clear that the constrained model performs with a  considerably higher consistency. The similarity threshold allows us to restrict adversarial movement during the training process and hence far fewer starting points need to be tried in order to identify a higher-performing classifier.

These numerical experiments have demonstrated the ability of the constrained pessimistic bilevel model to outperform the existing adversarial classifiers. In particular, we see improved performance over the classic approach to training a classifier and the existing pessimistic bilevel approach proposed by \cite{Brückner_Scheffer_2011}. Additionally, we see some improvement over the unconstrained bilevel model in \cite{benfield2024classificationstrategicadversarymanipulation}, although to  a lesser extent. However, we demonstrate a separate but important improvement over the unconstrained bilevel model in that the performance of the constrained model is considerably more consistent. The similarity threshold allows us to restrict how far the adversary can move their data and hence we can mitigate the risk of the adversary having too much freedom, leading to their data becoming unrealistic. Without this, the unconstrained model required trialling many starting points to identify a high-performing classifier. The constrained bilevel model, on the other hand, requires far fewer trials.

\section{Conclusion}
\label{sctn:Conclusion}
In this work, we introduced a novel approach to modelling text-based adversarial learning scenarios with pessimistic bilevel optimisation. This formulation sees the learner in upper-level attempt to construct a classifier that detects data produced by the adversary. Meanwhile, in the lower-level, the adversary modifies their data to evade detection. In particular, our model introduces a novel approach adversary that produces data that better reflects reality within a pessimistic bilevel model. We achieved this by introducing constraints to the lower-level that restrict the adversary from producing data that is too dissimilar from its original position. In this way to can help ensure that the data do not lose their intended message as the adversary modifies it in an attempt to evade detection by the learner. A current approach to the pessimistic model handles a similar concept by including a regularisation term that penalises on the Euclidean distance moved by the adversary. However, due to requiring a strongly convex lower-level objective function, they were restricted to only using the Euclidean norm to measure the extent to which the adversary's data changed. In this work, we make no assumptions on the convexity of the lower-level problem and hence we can utilise the cosine similarity as a measure of change.

For our initial numerical assessment, we compare the performance of our model to that of a baseline model which does not consider adversarial influence during the training process and see our model provide improved performance. We investigate the affect of varying the number of instances that the adversary may manipulate and find this to be a way of controlling the amount of influence the adversary has over the training process. A trade-off presents itself in that the adversary must be allowed enough data that we successfully capture the adversarial nature of the problem but not so much that the model becomes overly pessimistic. Finally, we compare the performance of our model with existing pessimistic bilevel approaches and see improved performance. In particular, due to the constraints on the adversary, we see our model train classifiers with a more consistently high performance compared to the existing methods.


\appendix
\newpage 
\section{Levenberg–Marquardt algorithm}
Below, we have a generic algorithm to find zeros of a mixed nonlinear
complementarity systems of the form that we have in \eqref{eqn:system}; see  \cite{LevenbergMarquardt} for more details on the algorithm and its convergence analysis:
\begin{algorithm}[H]
\caption{Global nonsmooth Levenberg–Marquardt method for mixed nonlinear
complementarity systems via Fischer–Burmeister function}\label{alg:LevenbergMarquardt}
\begin{algorithmic}[1]
\Require starting point $z^0 \in {\mathbb{R}^{q + mq}}$ and parameters $\kappa \in (0,1)$, $\epsilon > 0$, $\sigma, \beta \in (0,1)$, $\rho > 0$, $\gamma_1, \gamma_2 > 0$, $\eta \in (0,1)$ and $K \in \mathbb{N}$
\State set $k := 0$
\While{$||\Phi(z^k)|| \geq \epsilon$}
\State compute the derivative (or a numerical approximation) $-\nabla \Psi_{\text{FB}}(z^k)$
\State set $v_k := \min(\gamma_1, \gamma_2||\Phi(z^k)||)$ and compute $d^k$ as the uniquely determined solution of
\[(D_N \Phi(z^k)^T D_N \Phi(z^k) + v_k \mathbb{I}^p)d = -D_N \Phi(z^k)^T \Phi(z^k)\]

\If{$\Psi_{\text{FB}}(z^k + d^k) \leq \kappa \Psi_{\text{FB}}(z^k)$}
    \State set $z^{k+1} := z^k + d^k$
\Else

    \If{$\nabla \Psi_{\text{FB}}(z^k)d^k > -\rho ||\nabla \Psi_{\text{FB}}(z^k)|| ||d^k||$ or $||d^k|| < \rho_2$}
        \State set $d^k := -\nabla \Psi_{\text{FB}}(z^k)$
    \EndIf
    \State set $\alpha_k := \beta^{i_k}$ where $i_k \in \mathbb{N}$ is the smallest positive integer such that
    \[\Psi_{\text{FB}}(z^k + \beta^{i_k} d^k) \leq \Psi_{\text{FB}}(z^k) + \beta^{i_k} \sigma \nabla \Psi_{\text{FB}}(z^k) d^k\]
    \State $z^{k+1} := z^k + \alpha_k d^k$
\EndIf

\If{$ \frac{||\Phi(z^k)||}{||\Phi(z^{k-1})||} \geq \eta$ and $k > K$}
    \State \textbf{return} $z^k$
\EndIf

\State k:= k + 1
\EndWhile
\State \textbf{return} $z^k$
\end{algorithmic}
\end{algorithm}

\section{Leader and follower's derivatives under the logistic loss}

Let $x \in \mathbb{R}^q$ be the sample of data with corresponding label $y \in \{0,1\}$. Let $w \in \mathbb{R}^q$ be some weights of a prediction function $\sigma : \mathbb{R}^q \times \mathbb{R}^q \tto \mathbb{R}$ defined as,
\[\sigma(w, x) := \frac{1}{1+e^{-w^Tx}}.\]
We define the upper-level (learner's) loss function $\mathcal{L} : [0,1] \times \{0,1\} \rightarrow \mathbb{R}$ as,
\[\mathcal{L}(\sigma(w, x), y) = -y \log(\sigma(w, x)) - (1 - y) \log(1 - \sigma(w, x)).\]
Let $X \in \mathbb{R}^{m q}$ be the adversary's data with corresponding labels $Y \in \{0,1\}^m$. The upper-level objective $F : \mathbb{R}^q \times \mathbb{R}^{m q} \rightarrow \mathbb{R}$ is defined as,
\[F(w, X) := \frac{1}{n} \sum_{i=1}^n \mathcal{L}(\sigma(w,D_i),\gamma_i) + \frac{1}{m} \sum_{i=1}^m \mathcal{L}(\sigma(w,X_i),Y_i) + \frac{1}{\rho} \Vert w \Vert_2^2.\]
The derivative of the upper-level objective with respect to the classifier weights is given by,
\[\frac{\partial F}{\partial w} (w,X) = \frac{1}{n} D^T(\sigma(w, D) - \gamma) + \frac{1}{m} X^T(\sigma(w, X) - Y) + \frac{2}{\rho} w.\]
Let $\mathcal{D}_D = \text{diag}(\sigma(w, D))$ and $\mathcal{D}_{X} = \text{diag}(\sigma(w, X))$, where for some vector $v \in \mathbb{R}^n$, $\text{diag}(v)$ is the $n \times n$ diagonal matrix with diagonal elements $v$. The second derivative with respect to the learner's weights is then given by
\[\frac{\partial^2 F}{{ww}} (w, \theta) = \frac{1}{n} X \mathcal{D}_D(I_n - \mathcal{D}_D) D^T + \frac{1}{m} X \mathcal{D}_X(I_m - \mathcal{D}_X) X^T + \frac{2}{\rho},\]
where $I_n$ and $I_m$ are the $n\times n$ and $m\times m$ identity matrices respectively.

The derivative of the upper-level objective function with respect to the adversary's data is given by
\[\frac{\partial F}{\partial X_{ij}} (w,X) := \frac{1}{m} \sum_{k=1}^m \frac{\partial \mathcal{L}}{\partial X_{ij}} (\sigma(w,X_k),Y_k) = \frac{1}{m} \frac{\partial \mathcal{L}}{\partial X_{ij}} (\sigma(w,X_i),Y_i), \quad i = 1,\dots,m, \; j = 1,\dots,q,\]
where the derivative of the upper-level loss function with respect to the adversary's data is given by
\[\frac{\partial \mathcal{L}}{\partial X_{ij}} (w, X_i;Y_i) = w_j (\sigma(w, X_i) - Y_i), \quad i = 1,\dots,m, \; j = 1,\dots,q.\]
The second derivative of the upper-level objective with respect to the adversary's data is then given by
\[\frac{\partial^2 F}{\partial X_{i,j}X_{kc}} (w,X) = \frac{1}{m} \frac{\partial^2 \mathcal{L}}{X_{ij} X_{kc}} (\sigma(w,X_i), Y_i), \quad i,k = 1,\dots,m, \; j,c  = 1,\dots,q,\]
where the derivative of the upper-level loss function with respect to the adversary's data is given by the following cases,
\[\frac{\partial^2 \mathcal{L}}{\partial X_{ij} X_{kl}} (\sigma(w,X_i), Y_i) = \begin{cases}
    w_j^2 \sigma(w, X_i)(1 - \sigma(w, X_i)) & i = k, j = l \\
    w_j w_l \sigma(w, X_i)(1 - \sigma(w, X_i)) & i = k, j \neq l \\
    0 & i \neq j \neq k \neq l
\end{cases}.\]
The derivative of the loss with respect to both the classifier weights and the adversary's data is given by
\[\frac{\partial^2 \mathcal{L}}{\partial w_i X_{jk}} (\sigma(w,X_i), Y_i) = \begin{cases}
    X_{ji} w_k \sigma(w, X_j)(1 - \sigma(w, X_j)) + \sigma(w, X_j) - y_j & i = k \\
    X_{ji} w_k \sigma(w, X_j)(1 - \sigma(w, X_j)) & i \neq j \neq k
\end{cases}.\]
We define the lower-level (adversary's) objective as
\[f(w, X) := \sum_{i=1}^m \ell(\sigma(w,X_i),Y_i),\]
where $\ell: [0,1] \times \{0,1\} \tto \mathbb{R}$ is the lower-level loss function, defined  as the logistic loss with opposite class labels,
\[\ell(\sigma(w, x), y) := - (1 - y) \log(\sigma(w, x)) - y \log(1 - \sigma(w, x)).\]
The derivative of the lower-level objective with respect to the learner's weights is given by
\[\frac{\partial f}{\partial w} (w,X) = \frac{1}{m} X^T(\sigma(w, X) - (1 - Y)),\]
and the second derivative with respect the learner's weights is given by
\[\frac{\partial^2 f}{{ww}} (w, \theta) = \frac{1}{m} X \mathcal{D}_X(I_m - \mathcal{D}_X) X^T.\]
The derivative of the lower-level objective with respect to the adversary's data is given by
\[\frac{\partial f}{\partial X_{ij}} (w,X) = \frac{1}{m} \sum_{i=k}^m \frac{\partial \ell}{\partial X_{ij}} (\sigma(w, X_k), Y_k) = \frac{1}{m} \frac{\partial \ell}{\partial X_{ij}} (\sigma(w, X_k), Y_k), \quad i = 1,\dots,m, \; j = 1,\dots,q.\]
Note that the lower-level loss function can be expressed in terms of the upper-level loss function,
\[\ell(\sigma(w,x),y) = \mathcal{L}(\sigma(w,x),1-y).\]
Therefore, we can express the derivative of the lower-level objective function with respect to the adversary's data as follows,
\[\frac{\partial f}{\partial X_{ij}} (w, X) = \frac{1}{m} \frac{\partial \mathcal{L}}{\partial X_{ij}} (\sigma(w, X_i), 1-Y_i) \quad i = 1,\dots,m, \; j = 1,\dots,q.\]
The second derivative with respect to the adversary's data is then given as
\[\frac{\partial f}{\partial X_{ij} X_{kc}} (w, X) = \frac{1}{m} \frac{\partial \mathcal{L}}{\partial X_{ij} X_{jk}} (\sigma(w, X_i), 1-Y_i) \quad i,k = 1,\dots,m, \; j,c = 1,\dots,q.\]
Finally, the derivative with respect to the learner's weights and the adversary's data is given by
\[\frac{\partial f}{\partial w_i X_{jk}} (w, X) = \frac{1}{m} \frac{\partial \mathcal{L}}{\partial w_i X_{jk}} (\sigma(w, X_i), 1-Y_i) \quad i,j = 1,\dots,m, \; k = 1,\dots,q.\]


Let $X^0 \in \mathbb{R}^{m q}$ be the start point of the adversary's data. We define the constraints $g : \mathbb{R}^{m q} \rightarrow (-1,1)$ of the lower-level as
\[g(X) := \begin{pmatrix}
    g_1(X_1) \\
    \vdots \\
    g_m(X_m)
\end{pmatrix}\]
where each constrain function $g_i(X_i) : \mathbb{R}^q \rightarrow \mathbb{R}, \; i = 1,\dots,m$, is defined as,
\[
g_i(X) = \delta - d(X_i,X_i^0), \; i \in \{1,\dots,m\},
\]

where
\[d(X_i, X_i^0) = \frac{X_i \cdot X_i^0}{\Vert X_i \Vert \Vert X_i^0 \Vert}, \; i \in \{1,\dots,m\},\]
and $\delta \in \mathbb{R}$ is the similarity threshold. The derivative of the constraints with respect to the classifier weights is $0$. The derivative with respect to the adversary's data is given by the following cases,
\[\frac{\partial g_i(X)}{\partial X_{jk}} = \begin{cases}
    \frac{X_{ik}^0}{\Vert X_i \Vert \cdot \Vert X_i^0 \Vert} - d(X_i, X_i^0) \frac{X_{ik}}{\Vert X_i \Vert ^2} & i = j \\
    0 & i \neq j
\end{cases}.\]
The second derivative with respect to the adversary's data is given by the cases,
\[\frac{\partial g_i(X)}{X_{jk} X_{lc}} = \begin{cases}
    \frac{X_{ic} X_{ik}^0 + X_{ik} X_{ic}^0}{\Vert X_i \Vert ^ 3 \Vert X_i^0 \Vert} - \frac{3 X_{ik} X_{ic} d(X_i, X_i^0)}{\Vert X_i \Vert ^4} & i = j = l, k \neq c \\
    \frac{2 X_{ik} X_{ik}^0}{\Vert X_i \Vert ^ 3 \Vert X_i^0 \Vert} - \frac{3 X_{ik}^2 d(X_i, X_i^0)}{\Vert X_i \Vert ^4} + \frac{d(X_i, X_i^0)}{\Vert X_i \Vert ^2} & i = j = l, k = c \\
    0 & \text{otherwise}.
\end{cases}\]

\section{AM-GM inequality} \label{Chapter:AMGM}
The inequality of arithmetic and geometric means (AM-GM inequality), first proved by \cite{amgm}, states that for any non-negative pair $x,y \geq 0$, the arithmetic mean, given by $\frac{1}{2}(x+y)$, is greater than the geometric mean, given by $\sqrt{xy}$. The AM-GM inequality can be summarised as,
\[\frac{1}{2}(x+y) \geq \sqrt{xy} \; \forall \; x,y \geq 0.\]

Now consider two real numbers $u,v \in \mathbb{R}$. We can apply the AM-GM inequality to their squares since $u^2,v^2 \geq 0$. This then implies the following,
\begin{align*}
    \frac{1}{2}(u^2 + v^2) \geq \sqrt{u^2 v^2} & \implies \frac{1}{2}(u^2 + v^2) \geq |uv| \geq uv \\
    & \implies \frac{1}{2}(u^2 + v^2) \geq uv.
\end{align*}
The final implication can then be applied to Proposition \ref{prop:concavity}.

\section{Proofs}\label{Proofs}
\proof[Proof of Proposition 1]
Let $X^* \in \argmin_{X \in \mathbb{R}^{m q}} f(w, X), X^* \neq 0$. Choose any indices $i \in \{1,\dots,m\}$ and $j,k \in \{1,\dots,q\}$ such that $X^*_{ik} \neq 0$. Let $X^\prime$ be such that $X^\prime_{lr} = X^*_{lr}$ for all $l \{1,\dots, m\}$ and $j \in \{1,\dots, q\}\setminus\{j,k\}$ and set
\[X^\prime_{i,j} = X^*_{ij} + \frac{w_k}{w_j} X^*_{ik}, \; X^\prime_{ik} = 0.\]
Then, $X^\prime \neq X^*$ and $w_0 + w^TX^\prime_i = w_0 + w^T X^*_i \, \forall \, i \in \{1,\dots,m\}$. It then follows that
\begin{align*}
    w_0 + w^TX^\prime_i = w_0 + w^T X^*_i \, \forall \, i \in \{1,\dots,m\} & \implies \sigma(w, X^\prime_i) = \sigma(w, X^*_i) \, \forall \, i \in \{1,\dots,m\} \\
    & \implies \ell(w, X^\prime_i) = \ell(w, X^*_i) \, \forall \, i \in \{1,\dots,m\} \\
    & \implies f(w,X^\prime) = f(w,X^*).
\end{align*}
Set $\delta$ such that
\[\delta < \min \left ( \min\{d(X^*_i, X^0_i) : i \in \{1,\dots,m\}\}, \, \min\{d(X^\prime_i, X^0_i) : i \in \{1,\dots,m\}\} \right ). \]
It follows that $g(X^*) < 0$ and $g(X^\prime) < 0$ and so $X^* \in S(w)$ and $X^\prime \in S(w)$.
\endproof

\proof[Proof of Proposition 2]
Let $X \in \mathbb{R}^{mq}$ be such that $g(X) = 0$ and $X^0_i \cdot X_i < 0 \, \forall \, i \in \{1,\dots,m\}$. Let $X^\prime \in \mathbb{R}^{m q}, \, X^\prime \neq X$ be such that for each $i \in \{1,\dots,m\}$, $X_i^\prime$ is permutation of $X_i$ and note that for any $i \in \{1,\dots,m\}$, the following equalities hold:
\begin{itemize}
    \item $\sum_{j=1}^q X^\prime_{ij} = \sum_{j=1}^q X_{ij}$
    \item $\sum_{j=1}^q {X^\prime_{ij}}^2 = \sum_{j=1}^q X_{ij}^2$
    \item $\; X_i^0 \cdot X_i^\prime = \sum_{j=1}^q X^0_{ij} X^\prime_{ij} = \sum_{j=1}^q X^0_{ij} X_{ij} = X_i^0 \cdot X_i$.
\end{itemize}
It then follows that
    \[d(X^\prime_i, X^0_i) = \frac{X_i^0 \cdot X_i^\prime}{\Vert X_i^0 \Vert \Vert X_i^\prime \Vert} = \frac{X_i^0 \cdot X_i}{\Vert X_i^0 \Vert \Vert X_i \Vert} = d(X_i, X^0_i) = \delta, \: i = 1,\dots,m.\]
We now look at the similarity score for a point on the line segment connecting $X_i$ and $X^\prime_i$ for some $i \in \{1,\dots,m\}$,
    \begin{align*}
        d(\lambda X_i + (1-\lambda) X^\prime_i, X^0_i) & = \frac{X^0_i \cdot (\lambda X_i + (1-\lambda) X_i^\prime)}{\Vert X^0_i \Vert \Vert \lambda X_i + (1-\lambda) X_i^\prime \Vert} \\
        & = \frac{\lambda X^0_i \cdot X_i + (1-\lambda) X^0_i \cdot X_i^\prime}{\Vert X^0_i \Vert \Vert \lambda X_i + (1-\lambda) X^\prime_i \Vert} \\
        & = \frac{X^0_i \cdot X_i}{\Vert X^0_i \Vert \Vert \lambda X_i + (1-\lambda) X^\prime_i \Vert}.
    \end{align*}
    Focussing on the denominator of the above, observe that for any $i \in \{1,\dots,m\}$, we can expand the following term,
    \begin{align*}
        \Vert \lambda X_i + (1-\lambda) X^\prime_i \Vert & = \sqrt{\lambda^2 \sum_{j=1}^q X_{ij}^2 + 2 \lambda (1-\lambda) \sum_{j=1}^q X_{ij} X^\prime_{ij} + (1-\lambda)^2 \sum_{j=1}^q {X^\prime_{ij}}^2} \\
        & = \sqrt{\sum_{j=1}^q X_{ij}^2 + 2 \lambda (1-\lambda) \sum_{j=1}^q X_{ij} X^\prime_{ij} - 2 \lambda (1-\lambda) \sum_{i=1}^q X_{ij}^2}.
    \end{align*}
    From the inequality of arithmetic and geometric means (AM-GM inequality, see Appendix \ref{Chapter:AMGM}), 
    \[X_{ij} {X_{ij}^\prime} \leq \frac{1}{2}\left ( X_{ij}^2 + {X_{ij}^\prime}^2 \right ) \; \forall \; i \in \{1,\dots,m\}, j \in \{1,\dots,q\}\]
    with equality if and only if $X_{ij} = X^\prime_{ij}$. It therefore follows that for all $i \in \{1,\dots,m\}$,
    \[\sum_{j=1}^q X_{ij} {X_{ij}^\prime} \leq  \frac{1}{2} \sum_{j=1}^q \left ( X_{ij}^2 + {X_{ij}^\prime}^2 \right ) = \frac{1}{2} \left ( \sum_{j=1}^q X_{ij}^2 + \sum_{j=1}^q {X_{ij}^\prime}^2 \right ) = \frac{1}{2} \left ( \sum_{j=1}^q X_{ij}^2 + \sum_{j=1}^q {X_{ij}}^2 \right ) = \sum_{j=1}^q X_{ij}^2.\]
    It then follows that,
    \begin{align*}
        & 2 \lambda (1-\lambda) \sum_{j=1}^q X_{ij} X_{ij}^\prime \leq 2 \lambda (1-\lambda) \sum_{j=1}^q {X_{ij}}^2 \\
        \implies & 2 \lambda (1-\lambda) \sum_{j=1}^q X_{ij} X^\prime_{ij} - 2 \lambda (1-\lambda) \sum_{j=1}^q {X_{ij}}^2 \leq 0 \\
        \implies & \sum_{j=1}^q X_{ij}^2 +  2 \lambda (1-\lambda) \sum_{j=1}^q X_{ij} X^\prime_{ij} - 2 \lambda (1-\lambda) \sum_{j=1}^q {X_{ij}}^2 \leq \sum_{j=1}^q X_{ij}^2 \\
        \implies & \Vert \lambda X_i + (1-\lambda) X^\prime_i \Vert \leq \Vert X_i \Vert.
    \end{align*}
    Since $X^\prime \neq X$, it must hold that there exists some $i \in \{1,\dots,m\}$ such that $X_{ij} \neq X^\prime_{ij}$. Since the AM-GM inequality gives equality if and only if $X_{ij} = X^\prime_{ij}$, it must hold that,
    \[\Vert \lambda X_i + (1-\lambda) X^\prime_i \Vert < \Vert X_i \Vert.\]
    Therefore, since $X^0_i \cdot X_i < 0$, it must hold that
    \[d(\lambda X_i + (1-\lambda) X^\prime_i, X^0_i) = \frac{X^0_i \cdot X_i}{\Vert X^0_i \Vert \Vert \lambda X_i + (1-\lambda) X_i^\prime \Vert} < \frac{X_i^0 \cdot X_i}{\Vert X_i^0 \Vert \Vert X_i \Vert} = \delta.\]
    From here it follows that
    \begin{align*}
        d(\lambda X_i + (1-\lambda) X^\prime_i, X^0_i) < \delta & \implies \delta - d(\lambda X_i + (1-\lambda) X^\prime_i, X^0_i) > 0 \\
        & \implies g(\lambda X_i + (1-\lambda) X^\prime_i) > 0.
    \end{align*}
    Therefore, the point lying on the line segment connecting $X$ and $X^\prime$ is outside the set defined by the constraints.
\endproof

For the next result,  some preliminary background is necessary.

Assuming $\varphi_p$ is locally Lipschitz continuous, we define its Clarke subdiffential at the point $\bar w \in \mathbb{R}^q$ as
\[\partial \varphi_p(\bar{w}) := \{s \in \mathbb{R}^q \mid s^Td \leq \varphi_p^o (\bar{w};d) \, \forall \, d \in \mathbb{R}^q\},\]
    where $\varphi_p^o(\bar{w};d)$ is the generalized directional derivative of $\varphi_p$ at $\bar{w}$ in direction $d \in \mathbb{R}^q$, defined as
    \[\varphi_p^o(\bar{w};d) := \limsup_{w \rightarrow\bar{w}, h \downarrow 0} \frac{\varphi(w + hd) - \varphi(w)}{h}.\]
Let the set-valued mapping $S_p : \mathbb{R}^q \tto \mathbb{R}^{m q}$ be defined as
\[
\forall w \in \mathbb{R}^q : \quad S_p(w) := \argmin_{X \in \mathbb{R}^{m q}} \{-F(w, X) \, | \, X \in S(w) \}.
\]
We say that $S_p$ is inner semicontinuous at a point $(\bar{w}, \bar{X}) \in \gph S_p$ if for each convergent sequence $\{w^k\}_{k \in \mathbb{N}} \subset \mathbb{R}^q$ such that $w^k \rightarrow \bar{w}$, there exists a sequence $\{X^k\}_{k \in \mathbb{N}}$ such that $X^k \rightarrow \bar{X}$ which satisfies $X^k \in S_p(w^k)$ for all sufficiently large $k \in \mathbb{N}$.

\begin{theorem}
\label{thm:system}
Let $\bar w$ be a local optimal solution of \eqref{eqn:bilevel}--\eqref{eqn:lowerlevelset}, and let the following conditions hold:
\begin{itemize}
\item[(a)] $S_p$ is inner semicontinuous at $(\bar w, \bar X)$ for some $\bar X \in S_p(\bar w)$ such that $\bar X_i \neq 0$ for all $i=1, \ldots, m$. 
\item[(b)] The following lower-level Mangasarian-Fromovitz constraint qualification (LMFCQ) holds at $\bar X$:
\begin{equation}
   \left[\nabla g(\bar X)^\top \beta=0, \;\, \beta \geq 0, \;\, g(\bar X)\leq 0, \;\, \beta^\top g(\bar X)=0\right] \;\; \Longrightarrow\;\;  \beta =0.
\end{equation}
\item[(c)] The following set-valued mapping is calm at $(0, \bar w, \bar X)$:
\[
\Phi(\theta):=\left\{(w, X)\left|\; g(X)\leq 0,\;\; f(w, X) - \varphi(w) + \theta\leq 0\right.\right\},
\]
where $\varphi: \mathbb{R}^q \rightarrow \mathbb{X}^{m q}$ is the lower-level value function, defined as
    \[\varphi(w) := \min_{X \in \mathbb{R}^{m q}} f(w, X),\]
and where $\Phi: \mathbb{R} \rightarrow \mathbb{R}^q \times \mathbb{R}^{m q}$ is said to be calm at a fixed point $(\bar \theta, (\bar w, \bar X)) \in \gph \Phi$ if there exist constants $\epsilon, \delta, L > 0$ such that
    \[\forall \theta \in \mathbb{B}_\epsilon(\bar{\theta}), \, \forall (w, X) \in \Phi(\theta) \cap \mathbb{B}_\delta (\bar w, \bar X), \, \exists (\tilde{w}, \tilde{X}) \in \Phi(\bar \theta): \quad \lvert \lvert (w, X) - (\tilde w, \tilde X) \rvert \rvert \leq L \lvert \lvert \theta - \tilde \theta \rvert \rvert \]
    where $\mathbb{B}_\epsilon(\hat \theta)$ is the closed $\epsilon$-ball around $\bar \theta$.
\end{itemize}
There exist $(\lambda, \beta, \hat \beta)$ such that the conditions \eqref{KTpes6}--\eqref{KTpes8.1} hold. 
\end{theorem}

\proof[Proof of Theorem \ref{thm:system}]
    Start by observing that $\varphi_p(w) = -\varphi_{op}(w)$ for all $w$, where 
\[
\varphi_{\textup{op}}(w):= \min\{-F(w,X)\,|\,X\in S(w)\}.
\]
Since $S_p$ is inner semicontinuous at $(\bar w, \bar X)$, so is $S$ at the same point. Hence, the combination of assumptions (a) and (b) implies that the Lipschitz continuity of $\varphi$ around $\bar w$. Subsequently, $S$ is graph-closed around $(\bar w, \bar X)$. It therefore follows from \cite{mordukhovich2012variational} that 
\begin{equation}\label{eq:Sub_min}
    \partial \varphi_{\textup{op}} (\bar w) \subset \left\{w^*|\;\, (w^*, 0)\in -\nabla F(\bar w, \bar X) + N_{\text{gph} S}(\bar w, \bar X)\right\}.
\end{equation}
Note that 
\[
\text{gph} S:=\left\{(w, X)|\;\, g(X)\leq 0, \;\, f(w, X)-\varphi(w)\leq 0\right\}
\]
and under assumption (c), it holds that 
\begin{equation}\label{eq:NormalConeCal}
     N_{\text{gph} S}(\bar w, \bar X) \subset \left\{\left.\left.\left[\begin{array}{c}
      \lambda \left(\nabla_w f(\bar w, \bar X) -\partial \varphi(\bar w) \right) \\[1ex]
       \lambda \nabla_X f(\bar w, \bar X) + \partial g(\bar X)^\top \gamma 
 \end{array}\right]\right| \lambda \geq 0, \;\, \gamma \geq 0, \;\, g(\bar X)\leq 0, \;\, \gamma^\top g(\bar X)=0\right.\right\}.
\end{equation}
On the other hand, thanks to (a) and (b), it holds that 
\begin{equation}\label{eq:SubVarphi}
    \partial \varphi(\bar w) \subset \left\{\nabla_w f(\bar w, \bar X)\left|\;\begin{array}{r}
        \nabla_X f(\bar w,\bar X) + \partial g(\bar X)^\top\hat\beta \ni 0  \\[1ex]
         \beta \geq 0, \;\, g(\bar X)\leq 0, \;\, \beta^\top g(\bar X)=0
    \end{array}\right.\right\}.
\end{equation}
Combining this with \eqref{eq:NormalConeCal}, it follows that 
\begin{equation}\label{eq:NormalConeCal_1}
     N_{\text{gph} S}(\bar w, \bar X) \subset \left\{\left.\left.\left[\begin{array}{c}
     0\\[1ex]
       \lambda \nabla_X f(\bar w, \bar X) + \partial g(\bar X)^\top \gamma 
 \end{array}\right]\right| \begin{array}{r}
        \nabla_X f(\bar w,\bar X) + \partial g(\bar X)^\top\hat\beta \ni 0  \\[1ex]
         \beta \geq 0, \;\, g(\bar X)\leq 0, \;\, \beta^\top g(\bar X)=0\\[1ex]
         \lambda \geq 0, \;\, \gamma \geq 0, \;\, g(\bar X)\leq 0, \;\, \gamma^\top g(\bar X)=0
    \end{array} 
 \right.\right\}.
\end{equation}
Considering the fact $0\in N_{\text{gph} S}(\bar w, \bar X)$ with $\bar X\in S(\bar w)$ and assumption (b) holds, we can easily check that 
\[
D*S(\bar w|\bar X)(0) =\{0\}.
\]
This implies that $S$ is Lipschitz-like at $(\bar w, \bar X)$ by the Mordukhovich criterion. Hence, $\varphi_{\textup{op}}$ is Lipschitz continuous around $\bar w$, and so is $\varphi_p$. Therefore, $\bar w$ being a local optimal solution of problem \eqref{eqn:bilevel}--\eqref{eqn:lowerlevelset}, in the sense of \eqref{eq:OptimalConcept}, it holds that  
\begin{equation}\label{eq:OptCond}
    0\in \partial \varphi_p(\bar w).
\end{equation}
Given that  $\partial \varphi_p(\bar w) =  \partial (-\varphi_p)(\bar w) =  -\partial \varphi_p(\bar w)$, considering the Lipschitz continuity of $\varphi_p$ and the corresponding property from the Clarke subdifferential concept, the result follows from the combination of the inclusion in equations \eqref{eq:Sub_min}, \eqref{eq:NormalConeCal}, and \eqref{eq:OptCond}. 
\endproof
\end{document}